\documentclass[sigconf,nonacm]{acmart}

\usepackage{booktabs}
\usepackage{graphicx}

\usepackage{amsmath,amssymb}
\usepackage{multirow}
\usepackage{xcolor}
\usepackage{listings}
\usepackage{subcaption}
\usepackage{tikz}
\usepackage{pgfplots}
\usepackage{placeins}
\usepackage{float}
\usepackage{caption}
\pgfplotsset{compat=1.18}

\usetikzlibrary{shapes.geometric, arrows.meta, positioning, calc, fit, backgrounds, patterns, decorations.markings, shadows, matrix}

\definecolor{awblue}{HTML}{1565C0}
\definecolor{awnavy}{HTML}{0D2137}
\definecolor{awlight}{HTML}{42A5F5}
\definecolor{awgreen}{HTML}{2E7D32}
\definecolor{awteal}{HTML}{00838F}
\definecolor{aworange}{HTML}{E65100}
\definecolor{awred}{HTML}{C62828}
\definecolor{awgray}{HTML}{607D8B}
\definecolor{awbg}{HTML}{F5F7FA}
\definecolor{awpurple}{HTML}{6A1B9A}
\definecolor{awamber}{HTML}{FF8F00}

\begin{document}

\fancyhead[RE]{}
\fancyhead[LE]{}
\fancyhead[RO]{}
\fancyhead[LO]{}
\renewcommand{\shortauthors}{}

\title{AgentWorld: Personality-Aware Reliability Evaluation \\ for Agentic Information Retrieval}

\author{Gunja Agarwal}
\affiliation{\institution{}\country{}}
\email{guagarwal@paypal.com}

\author{Arup Kumar Das}
\affiliation{\institution{}\country{}}
\email{arupdas@paypal.com}

\author{Arun Menon}
\affiliation{\institution{}\country{}}
\email{arumenon@paypal.com}

\author{Jitesh Chandra Mishra}
\affiliation{\institution{}\country{}}
\email{jimishra@paypal.com}

\author{Vignesh Divakaran}
\affiliation{\institution{}\country{}}
\email{vignd@paypal.com}

\begin{abstract}
Evaluation of agentic information retrieval remains limited to scripted interactions with uniform users, missing both natural personality diversity and adversarial brittleness. We present \textbf{AgentWorld}, a simulation framework combining (i)~Big Five (OCEAN) personality-driven user populations with stateful tool-use environments; (ii)~the pass$^k$ consistency metric with structured fault classification, partial-credit scoring, and dual-control handoff verification; (iii)~score-thresholded training-data export in six fine-tuning formats; and (iv)~an adversarial Risk Analyser that snapshots required-intermediate-state spines, branches Monte-Carlo rollouts under four task-aware perturbation types, and quantifies risk via $\Delta P / \Delta T$ scoring, Dempster--Shafer evidence fusion, and Shapley attack-category attribution. Three experiments demonstrate the framework: a conversational analytics agent across 10 OCEAN personas (240 evaluator judgments); a customer-support agent across 5 tasks $\times$ 4 persona variants (19 simulations); and adversarial stress-testing of 5 tasks revealing pre-existing trajectory brittleness ($V_{\min}=0.375$ without perturbation) and tool/infrastructure-layer attack dominance (Shapley: 46\% system, 38\% action). Personality variation surfaces failure modes uniform testing cannot expose---cross-domain leakage, contextual drift, a 0.27-point quality gap, and 50\% vs.\ 100\% pass-rate across personas on the same task---while the Risk Analyser quantifies trajectory-level brittleness that pass$^k$ alone cannot measure.
\end{abstract}

\ccsdesc[500]{Information systems~Evaluation of retrieval results}
\ccsdesc[500]{Information systems~Users and interactive retrieval}
\ccsdesc[500]{Computing methodologies~Multi-agent systems}
\ccsdesc[300]{Computing methodologies~Agent / discrete models}
\ccsdesc[300]{Security and privacy~Software security engineering}

\keywords{agent evaluation, agentic information retrieval, personality modeling, user simulation, reliability metrics, conversational AI, OCEAN Big Five, tool-use evaluation, multi-agent topology}

\maketitle

\begin{figure*}[!t]
    \centering
    \includegraphics[width=\textwidth]{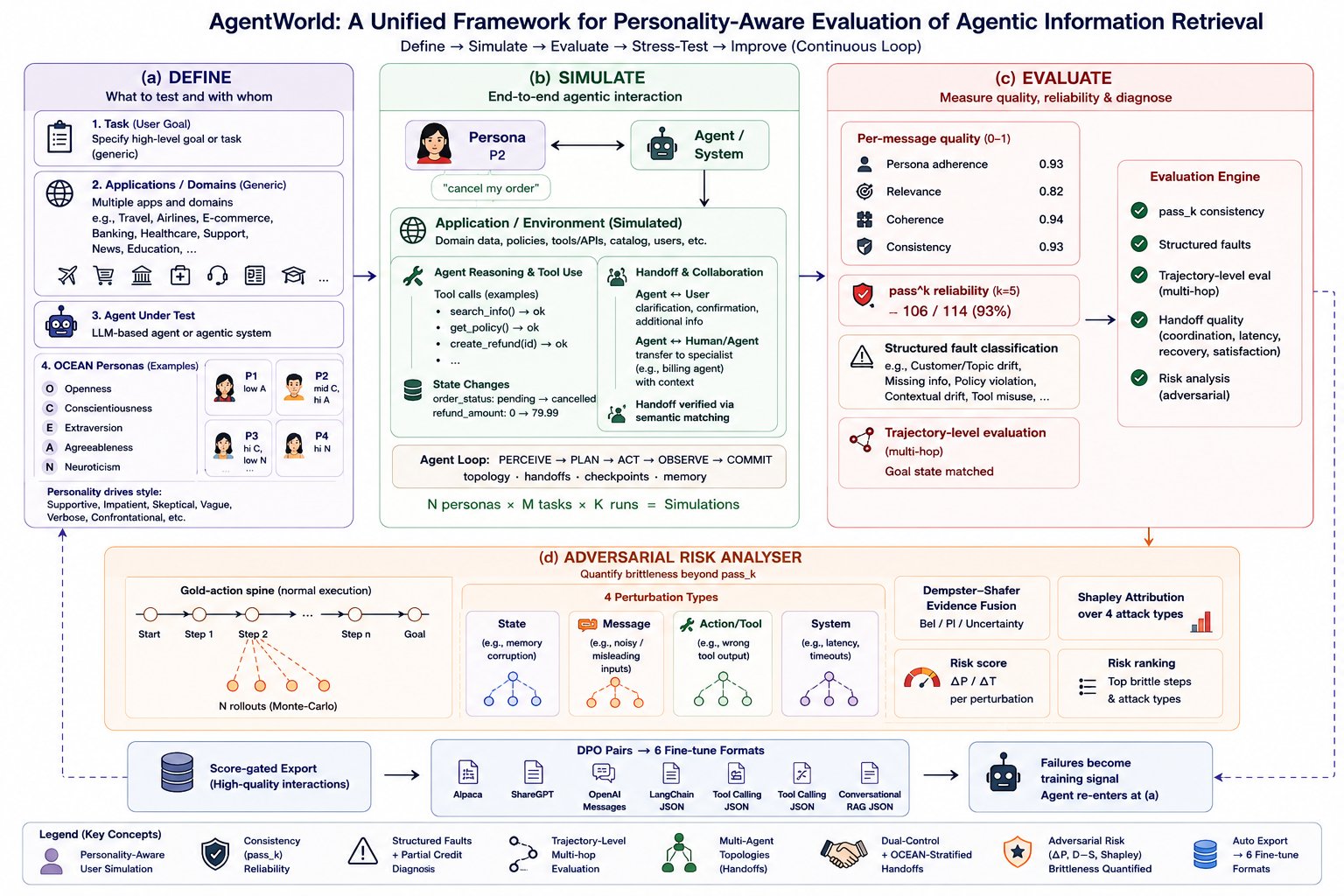}
    \caption{\textbf{AgentWorld: a unified framework for personality-aware evaluation of agentic information retrieval.} Four phases run in a continuous \emph{Define $\rightarrow$ Simulate $\rightarrow$ Evaluate $\rightarrow$ Stress-Test $\rightarrow$ Improve} loop. \textbf{(a)~Define}: user goal, application/domain, agent under test, OCEAN personas. \textbf{(b)~Simulate}: end-to-end multi-turn interaction with tool use, handoffs, and state changes under a \textsc{Perceive}--\textsc{Act}--\textsc{Commit} cycle. \textbf{(c)~Evaluate}: per-message behavioural scores, pass$^k$ reliability, structured fault classification, and trajectory- and handoff-level evaluation. \textbf{(d)~Adversarial Risk Analyser} (\S\ref{sec:riskanalyser}, \S\ref{sec:quantrisk}): required intermediate-state spine with Monte-Carlo branching, four task-aware perturbation types, Dempster--Shafer evidence fusion, and Shapley attribution. Score-gated export closes the loop---failures become training signal.}
    \label{fig:framework}
\end{figure*}

\section{Introduction}

AI agents that interact with users through natural language have become central to information retrieval systems---from conversational recommendation engines to agentic retrieval systems that orchestrate multi-step tool calls to fulfill complex queries~\cite{yao2023react}. Despite rapid advances in agent capabilities, evaluation remains a bottleneck: most agents are tested against scripted scenarios that fail to capture the diversity, unpredictability, and emotional range of real users. An agent that retrieves information perfectly for a polite, well-formed query may fail when a frustrated user asks the same question indirectly, switches topics mid-conversation, or demands immediate action.

We identify five gaps in how AI agents are currently tested:

\begin{enumerate}
    \item \textbf{No personality coverage.} Real users come in all types---patient, impatient, confrontational, vague, sarcastic. But test suites typically only use polite, well-structured queries.
    \item \textbf{No consistency measurement.} An agent that passes a test once might fail three times out of five. Testing once and calling it ``done'' gives a false sense of reliability.
    \item \textbf{No detailed failure diagnosis.} When something goes wrong, teams need to know \textit{what} failed and \textit{why}---not just a pass/fail label.
    \item \textbf{No path from testing to fixing.} Problems found during evaluation have to be manually turned into training data. Testing and improvement are disconnected.
    \item \textbf{No adversarial brittleness measurement.} Agents are tested against cooperative users; we do not know how much each layer of a multi-turn trajectory (data, communication, tool, infrastructure) erodes task success when an external adversary intervenes, nor how to attribute aggregate risk across attack categories with calibrated~uncertainty.
\end{enumerate}

AgentWorld (Figure~\ref{fig:framework}) is a simulation and evaluation framework that addresses these five gaps, together with three extensions specific to agentic IR (trajectory-level evaluation across multi-hop tool calls, $N$-agent topologies for multi-stage retrieval pipelines, and OCEAN-stratified handoff verification). Our contributions are:

\begin{itemize}
    \item \textbf{A four-phase closed-loop pipeline} (\textit{Simulate $\rightarrow$ Inject $\rightarrow$ Evaluate $\rightarrow$ Export}) combining OCEAN personality-driven user simulation with stateful tool-use environments and the pass$^k$ consistency metric---a combination absent from prior frameworks (Table~\ref{tab:comparison}; \S\ref{sec:framework}, \S\ref{sec:persona}, \S\ref{sec:apps}).

    \item \textbf{Structured fault classification, partial-credit scoring, and OCEAN-stratified handoff verification} that replace binary pass/fail with a two-part diagnosis (\textit{who} failed, \textit{what} went wrong), award partial credit for near-misses, and extend $\tau^2$-bench's dual-control model with a 4-step handoff cycle, three coordination metrics, and semantic-matching verification---revealing which personality profiles trigger which failures (\S\ref{sec:faults}, \S\ref{sec:apps}).

    \item \textbf{Score-thresholded export and personality-aware model comparison.} Behavioural scores gate which conversations enter six fine-tuning formats (including DPO pairs), and a model-comparison runner tracks performance across OCEAN quadrants and checkpoints, catching personality-specific regressions (\S\ref{sec:export}, \S\ref{sec:modelcomp}).

    \item \textbf{Adversarial Risk Analyser}---an external-adversary layer that generates four classes of task-aware perturbations (state, message, action, system), branches Monte-Carlo rollouts from required intermediate states to estimate $V(s_t)$, and produces per-perturbation Risk Scores ($\Delta P / \Delta T$), Dempster--Shafer belief/plausibility/uncertainty, and Shapley attribution across attack categories (\S\ref{sec:riskanalyser}, \S\ref{sec:quantrisk}).
\end{itemize}
\section{Related Work}

\begin{sloppypar}
\textbf{Benchmarks and user simulation.}
$\tau$-bench~\cite{yao2024tau} introduced consistency measurement (pass$^k$) for customer-service agents; $\tau^2$-bench~\cite{yao2025tau2} added dual-control coordination; AppWorld~\cite{trivedi2024appworld} provided stateful apps with 457~APIs; SAGE~\cite{sage2025} and TED~\cite{ted2026} added knowledge-grounded users and automated error analysis; WebArena~\cite{zhou2024webarena}, OSWorld~\cite{xie2024osworld}, AgentBench~\cite{liu2024agentbench}, and BFCL~\cite{bfcl2024} test agents in browsers, operating systems, and function-calling settings.
On the user-simulation side, Generative Agents~\cite{park2023generative} demonstrated memory-driven LLM characters; SOTOPIA~\cite{sotopia2024,sotopias4_2025} evaluates social intelligence with personality-like traits; and LLMs can be reliably steered to Big Five profiles~\cite{jiang2024evaluating}. TinyTroupe~\cite{tinytroupe2024}, CAMEL~\cite{li2023camel}, and AutoGen~\cite{wu2023autogen} support multi-agent persona simulation.
None of the surveyed frameworks combine OCEAN personality, stateful apps, pass$^k$, structured faults, training export, and adversarial risk in a single pipeline (Table~\ref{tab:comparison}).
\end{sloppypar}

\textbf{Adversarial robustness and risk attribution.} AgentDojo~\cite{agentdojo2024}, InjecAgent~\cite{injecagent2024}, AgentPoison~\cite{agentpoison2024}, and ToolEmu~\cite{toolemu2024} attack agents at the prompt or tool surface and report binary pass/fail. Our Risk Analyser instead measures impact at the trajectory level via counterfactual snapshot-and-branch, quantifying how much each attack reduces $V(s_t)$ over a multi-step plan. We fuse heterogeneous evidence using Dempster--Shafer belief functions~\cite{shafer1976,yager1987} with explicit ignorance modelling, attribute aggregate risk via exact Shapley values~\cite{shapley1953} over four attack categories, and blend finite-difference sensitivity with counterfactual causal impact inspired by chain-of-thought step-importance methods~\cite{madaan2022}. Attack graphs~\cite{sheyner2002} are left to future work.

\begin{table}[t]
\caption{Feature support across surveyed agent-evaluation frameworks (\checkmark = supported, --~=~not supported, Part.~=~partial). Capabilities are drawn from each framework's documented features as of submission.}
\label{tab:comparison}
\setlength{\tabcolsep}{2pt}
\fontsize{6}{7.5}\selectfont
\begin{tabular}{@{}l c c c c c c c@{}}
\toprule
\textbf{Capability} & \rotatebox{70}{\textbf{AgentWorld}} & \rotatebox{70}{$\tau$\textbf{-bench}} & \rotatebox{70}{$\tau^2$\textbf{-bench}} & \rotatebox{70}{\textbf{AppWorld}} & \rotatebox{70}{\textbf{SOTOPIA}} & \rotatebox{70}{\textbf{SAGE}} & \rotatebox{70}{\textbf{TED}} \\
\midrule
$N$-agent topologies & \checkmark & -- & -- & -- & \checkmark & -- & -- \\
OCEAN personality & \checkmark & -- & -- & -- & Part. & -- & -- \\
Stateful simulated apps & \checkmark & \checkmark & \checkmark & \checkmark & -- & -- & -- \\
pass$^k$ reliability & \checkmark & \checkmark & \checkmark & -- & -- & -- & -- \\
Structured fault classif. & \checkmark & -- & -- & -- & -- & -- & -- \\
Partial-credit scoring & \checkmark & -- & -- & -- & -- & -- & -- \\
Dual-control eval & \checkmark & -- & \checkmark & -- & -- & -- & -- \\
HTTP-only injection & \checkmark & -- & -- & -- & -- & -- & -- \\
Training export (6 fmt) & \checkmark & -- & -- & -- & -- & -- & -- \\
Model comparison & \checkmark & -- & -- & -- & -- & -- & -- \\
100+ LLM providers & \checkmark & -- & -- & -- & \checkmark & -- & -- \\
Gymnasium wrapper & \checkmark & -- & -- & -- & -- & -- & -- \\
Adversarial perturbation eval & \checkmark & -- & -- & -- & -- & -- & -- \\
Risk quant.\ ($\Delta P$, D--S, Shapley) & \checkmark & -- & -- & -- & -- & -- & -- \\
\bottomrule
\end{tabular}
\end{table}

\section{AgentWorld Framework}
\label{sec:framework}

\textbf{A worked example, in 30 seconds.} You point AgentWorld at your agent's HTTP endpoint, pick a task (e.g., \textit{``cancel a pending payment''}) and four user personas with different OCEAN profiles---including a confrontational one ($t_A=0.3, t_N=0.7$) and a polite-but-distractable one ($t_C=0.4, t_A=0.7$)---and click run. AgentWorld simulates four parallel conversations, scores every message on coherence and relevance, checks whether the cancellation actually happened in the app's database, classifies failures (\textit{Agent/Wrong-parameters}, \textit{Agent/Missing-confirmation}), and exports the transcripts as JSONL fine-tuning data. You see that your agent works for 3 of 4 personas but over-complies with the polite-distractable user---a failure mode self-testing would never have surfaced.

AgentWorld's architecture follows a four-phase pipeline (Figure~\ref{fig:framework}): \textit{simulate} a personality-driven user, \textit{inject} the user's messages into the agent under test, \textit{evaluate} the conversation on multiple axes, and \textit{export} the results as training data---with a feedback loop that re-runs improved agents on the same evaluation pack.

\subsection{Personality-Driven User Simulation}
\label{sec:persona}

The core idea behind AgentWorld is that real users have different personalities, and an agent that works for one personality type may fail for another. To capture this, each simulated user is defined by five personality scores based on the well-established Big Five (OCEAN) model from psychology~\cite{costa1992revised,goldberg1990}:
\begin{equation}
\mathbf{t} = (t_O, t_C, t_E, t_A, t_N) \in [0, 1]^5
\end{equation}
where $\mathbf{t}$ is the personality trait vector, and each component is a score between 0 (low) and 1 (high) for one of the five dimensions: $t_O$ = Openness (curiosity, creativity), $t_C$ = Conscientiousness (organisation, discipline), $t_E$ = Extraversion (talkativeness, energy), $t_A$ = Agreeableness (cooperation, politeness), and $t_N$ = Neuroticism (emotional volatility, anxiety).

For example, a user with $t_A = 0.1$ (very low Agreeableness) and $t_N = 0.9$ (very high Neuroticism) is confrontational and emotionally reactive---the kind of user who stresses an agent with complaints and demands. A user with $t_A = 0.9$ is polite and accommodating, making it easy for the agent to succeed (but potentially too easy, hiding real problems).
Teams can also add custom traits relevant to their domain:
\begin{equation}
\mathbf{t}_{\text{ext}} = \mathbf{t} \oplus \{(c_i, w_i)\}_{i=1}^{m}
\end{equation}
where $\mathbf{t}_{\text{ext}}$ is the extended trait vector, $\oplus$ denotes concatenation, $c_i \in [0,1]$ is the value of the $i$-th custom trait (e.g., tech-savviness = 0.8, patience = 0.2), $w_i$ is its weight in similarity calculations, and $m$ is the total number of custom traits added.

\begin{figure}[t]
    \centering
    \includegraphics[width=\columnwidth]{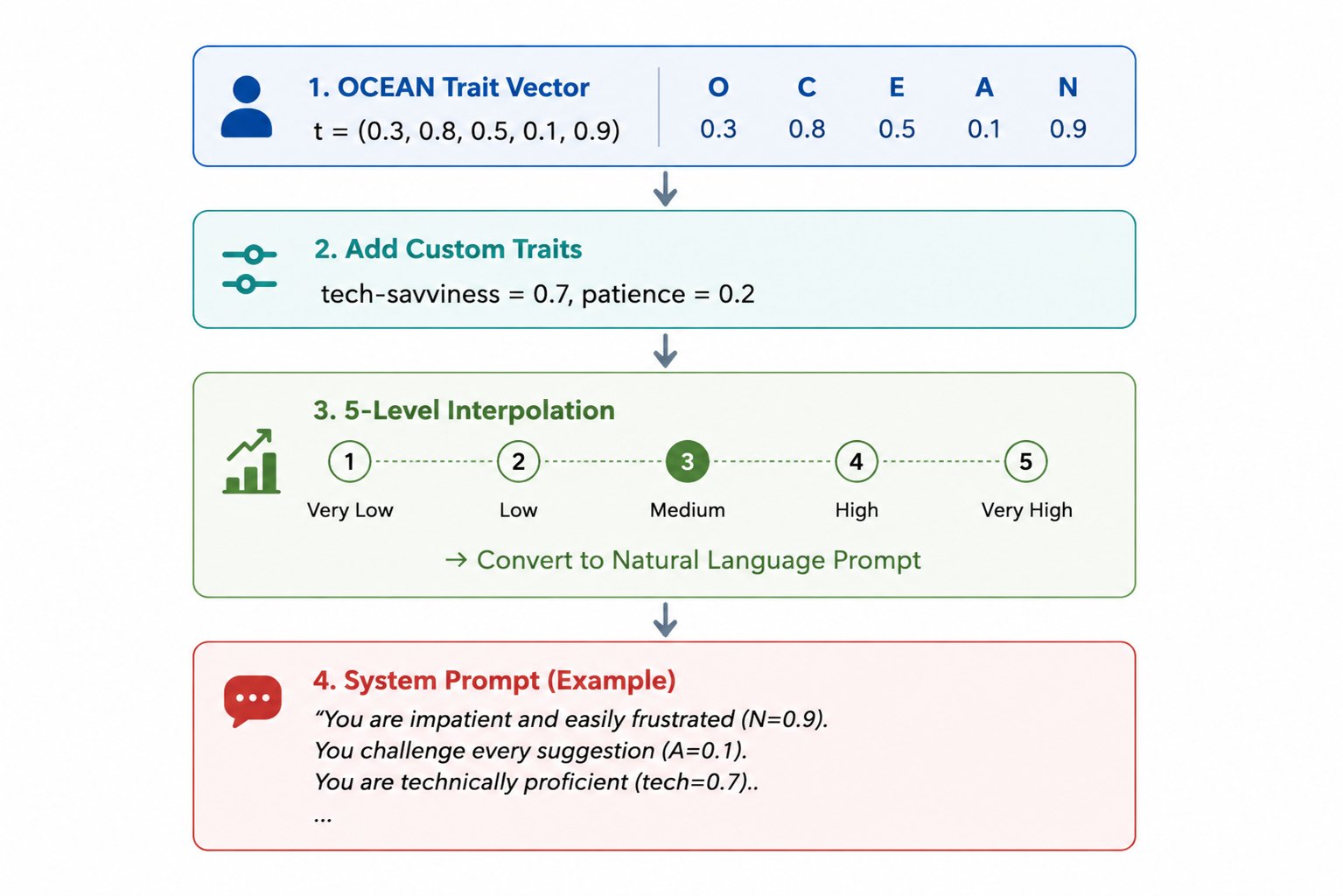}
    \caption{Persona generation pipeline. (1) A continuous OCEAN trait vector defines the personality. (2) Optional custom traits (e.g., tech-savviness, patience) are appended. (3) Each score is interpolated across five levels (Very Low to Very High). (4) The result is a natural-language system prompt that governs simulated user behaviour.}
    \label{fig:persona}
\end{figure}

\textbf{How traits become behaviour.} Each personality score is converted into a natural-language instruction using a 5-level mapping (Figure~\ref{fig:persona}). For example, Agreeableness below 0.2 maps to ``competitive, skeptical, challenging'' while above 0.8 maps to ``highly accommodating, conflict-averse.'' A user with Neuroticism above 0.8 and Agreeableness below 0.2 will generate confrontational, emotional messages---exactly the type of interaction that conventional test suites never cover.

\begin{sloppypar}
\textbf{Generating user populations at scale.} For large evaluations, AgentWorld can automatically generate hundreds of users by sampling personality scores from configurable distributions: $t_j \sim \mathcal{N}(\mu_j, \sigma_j^2)$, clamped to~$[0,1]$, where~$t_j$ is the score for dimension~$j$, $\mu_j$~is the desired mean (e.g., $\mu_A\!=\!0.2$ for a population of disagreeable users), and $\sigma_j^2$~controls how much variation there is around that mean.
This means teams can create, for example, a population skewed toward impatient users to stress-test agent resilience. To measure how similar two personas are, AgentWorld uses weighted cosine similarity:
\end{sloppypar}
\begin{multline}
\text{sim}(\mathbf{t}_a, \mathbf{t}_b) = \frac{\sum_j w_j^2 \, t_{a,j} \, t_{b,j}}{\|\mathbf{w} \odot \mathbf{t}_a\| \cdot \|\mathbf{w} \odot \mathbf{t}_b\|}\\
    = \frac{(\mathbf{w}\odot\mathbf{t}_a)\cdot(\mathbf{w}\odot\mathbf{t}_b)}{\|\mathbf{w} \odot \mathbf{t}_a\| \, \|\mathbf{w} \odot \mathbf{t}_b\|}
\end{multline}
where $\mathbf{t}_a$ and $\mathbf{t}_b$ are two persona trait vectors, $w_j$ is the weight assigned to dimension $j$ (allowing teams to emphasise certain traits over others), $t_{a,j}$ is persona $a$'s score on dimension $j$, and $\odot$ denotes element-wise multiplication. This is the standard cosine similarity between the weighted vectors $\mathbf{w} \odot \mathbf{t}_a$ and $\mathbf{w} \odot \mathbf{t}_b$, so the weights $w_j$ enter numerator and denominator consistently. A similarity of 1.0 means the two personas are identical; values near 0 mean they are very different. This enables clustering personas into groups and measuring diversity coverage across the population.

\textbf{Why this matters.} Most testing tools vary the \textit{task}---they try different questions and scenarios. AgentWorld also varies the \textit{user}. The same task (``transfer \$50 to my friend'') produces fundamentally different conversations when attempted by a patient user versus an impatient, skeptical one---and exposes completely different failure modes.

\subsection{Communication Topologies for Multi-Agent IR}

In agentic IR, queries often flow through multiple agents---a front-line assistant routes to a domain specialist, who may escalate to a supervisor. AgentWorld makes this structure explicit and configurable by assigning a network topology to each simulation, implemented using NetworkX. Five built-in topologies are supported: \textit{full mesh} (collaborative retrieval), \textit{hub-spoke} (centralised intent routing), \textit{hierarchical} (tiered escalation), \textit{small-world} (local clusters with long-range shortcuts), and \textit{scale-free} (one highly connected hub with peripheral specialists). By controlling topology per simulation, teams can test whether their multi-agent retrieval system performs better with centralised routing or distributed collaboration.

\subsection{Simulated Application Environments}
\label{sec:apps}

In real deployments, agents don't just chat---they retrieve account information, process transactions, check order status, and modify settings. AgentWorld simulates these stateful, tool-rich environments so that the \textit{full retrieval-and-action chain} can be evaluated, not just the conversational surface. Each simulated app exposes four primitives---\textit{actions} (typed-parameter tool calls), \textit{state} (persistent data such as balances and order history), \textit{observations} (notifications pushed to the agent), and an immutable \textit{audit trail} for replay and verification.

\textbf{Execution cycle.} Each turn follows an atomic \textsc{Perceive} $\rightarrow$ \textsc{Act} $\rightarrow$ \textsc{Commit} cycle: the agent observes app state and notifications, emits a tool call, and the engine validates parameters and executes the action transactionally, recording a state diff. Simulations are reproducible (seeded), pausable, and stream live events to the dashboard.

\begin{sloppypar}
\textbf{Extensibility and built-ins.} Apps can be added in two ways: \textit{(a)}~a Python class implementing the protocol, or \textit{(b)}~pure JSON via a declarative logic engine (\texttt{validate}, \texttt{set\_state}, \texttt{conditional})---letting domain experts ship test environments without writing code.
AgentWorld ships built-in apps across six categories (Payment, Shopping, Communication, Calendar, Social, Custom), including a customer-support app with dozens of tools plus Airlines, Retail, and loyalty domains. Three roles (\textit{Peer}, \textit{Service Agent}, \textit{Customer}) carry role-specific permissions following the $\tau^2$-bench model~\cite{yao2025tau2}.
\end{sloppypar}

\begin{figure*}[!t]
    \centering
    \includegraphics[width=\textwidth]{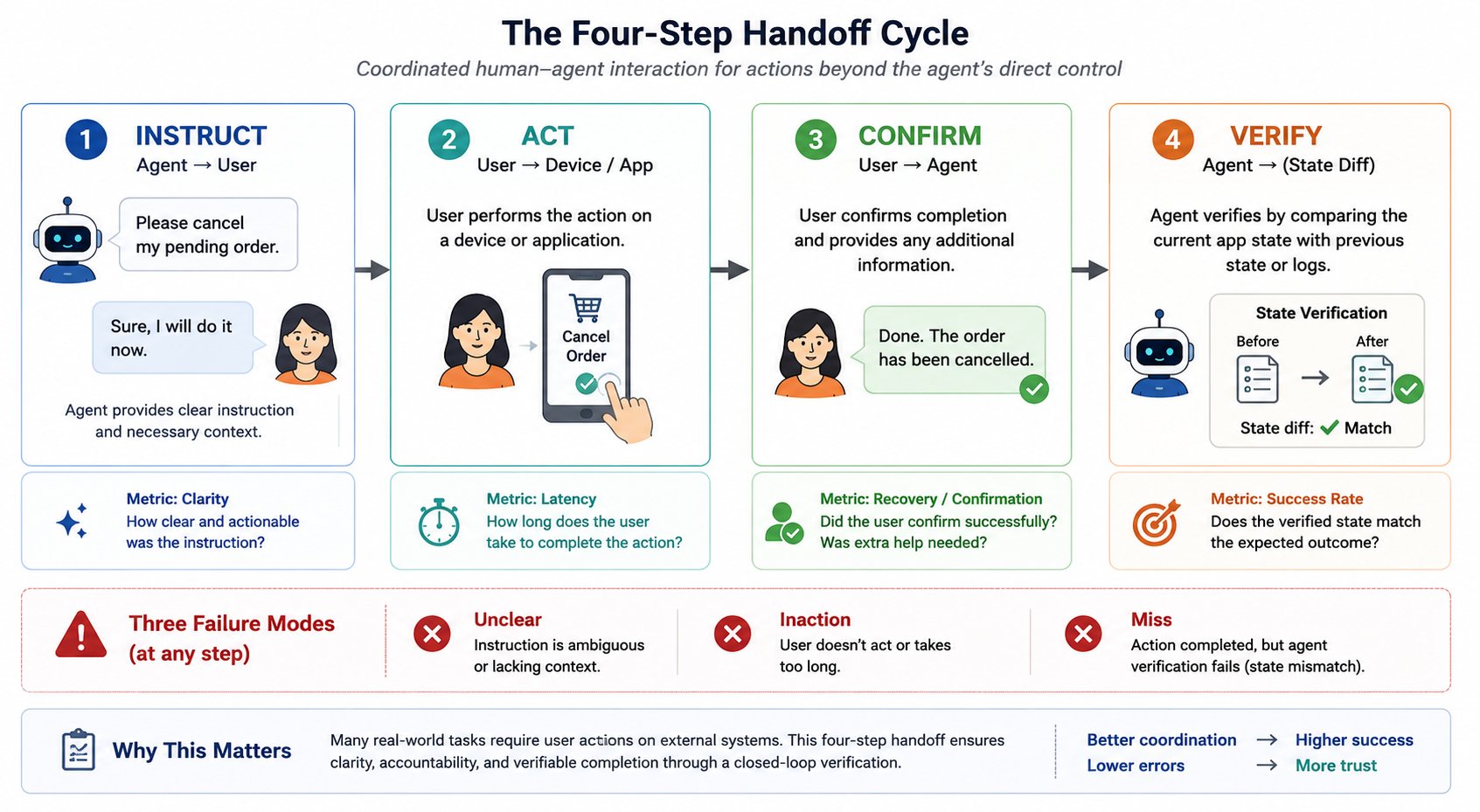}
    \caption{\textbf{The four-step handoff cycle.} Coordinated human--agent interaction for actions beyond the agent's direct control. (1)~\textsc{Instruct}: the agent provides clear instructions and context. (2)~\textsc{Act}: the user performs the action on a device or application. (3)~\textsc{Confirm}: the user confirms completion and provides any additional information. (4)~\textsc{Verify}: the agent verifies by comparing the current app state with the previous state or logs. Three failure modes---\textit{unclear}, \textit{inaction}, and \textit{miss}---and four metrics (clarity, latency, recovery/confirmation, success rate) are tracked at each step.}
    \label{fig:handoff}
\end{figure*}

\textbf{Tasks: the unit of evaluation.} A \textit{task} bundles a goal with success criteria. AgentWorld supports two task types corresponding to how agents are deployed in practice:
\begin{itemize}
    \setlength{\itemsep}{0pt}\setlength{\parskip}{0pt}\setlength{\topsep}{1pt}
    \item \texttt{TaskDefinition} (\textit{single-actor}): the agent has full tool access and works alone, e.g., \textit{``fetch the last five transactions.''} Measures raw capability.
    \item \texttt{DualControlTaskDefinition} (\textit{coordination}): tool access is \textit{split}---the agent controls backend tools (account DB, transaction APIs); the user controls a device the agent cannot reach (settings, 2FA app, hardware). Success requires a \textbf{handoff} (Figure~\ref{fig:handoff}): the agent instructs, the user acts, the agent verifies.
\end{itemize}

\textbf{Task lifecycle.} Each task flows through five stages---\textit{Define} (goal, domain), \textit{Configure} (apps, states, personas), \textit{Execute} ($k$ trials), \textit{Verify} (final state vs.\ goal conditions), \textit{Analyze} (pass$^k$ + fault tags)---and ends when all goal conditions are met (or a step limit is reached).

\textbf{Goal conditions} (one or more per task; all must pass): \textit{state} (e.g., \textit{balance=\$450}), \textit{action} (e.g., \textit{must call \texttt{issue\_refund}}), \textit{handoff} (did the user complete their part, verified via keyword + semantic match), \textit{output} (e.g., \textit{must mention refund policy}), and \textit{replay} (final state matches a known-good run, for regression testing).

\textbf{Agent memory.} Each simulated agent has two types of memory, inspired by Generative Agents~\cite{park2023generative}: \textit{episodic} (a log of what happened) and \textit{semantic} (higher-level insights derived by reflecting on recent events). Retrieval balances relevance, recency, and importance, letting agents develop coherent long-term understanding across extended conversations.

\textbf{HTTP injection of external agents.}\label{sec:injection} An agent exposing an HTTP endpoint can be evaluated without modification: the injection layer routes simulation messages to the external endpoint and integrates the response as a native agent turn. Three privacy tiers (Minimal: ID + hash; Basic: + traits; Full: + background) control how much persona data is exposed; failures are handled via circuit breaker, exponential backoff, and P50/P99 latency tracking.

\subsection{Adversarial Risk Analyser}
\label{sec:riskanalyser}

The Risk Analyser is an adversarial layer on top of AgentWorld that quantifies how much each layer of the simulation erodes task success when an external adversary intervenes. Four innovations distinguish it from prior agent red-teaming:

\begin{itemize}
    \setlength{\itemsep}{2pt}\setlength{\parskip}{0pt}\setlength{\topsep}{1pt}
    \item \textbf{Layer-aligned perturbation catalogue.} One perturbation type per AgentWorld layer---data, communication, tool, infrastructure (Table~\ref{tab:perturbations})---rather than prompt-only attacks at the input surface.
    \item \textbf{Task-aware LLM-generated perturbations.} Adversarial actions are synthesised at task-selection time and grounded in the selected task's required intermediate-state sequence, app identifiers, and state schema---producing executable perturbations rather than static templates. The user can review, edit, or regenerate the set; the final set is persisted with the analysis profile for reproducibility.
    \item \textbf{Trajectory-level $\Delta P$ via snapshot-and-branch.} Monte-Carlo branching from required intermediate states estimates $V(s_t)$ and $V'(s_t, p)$, measuring the actual reduction in success probability over the multi-step plan rather than a binary attack-success flag.
    \item \textbf{Uncertainty-aware aggregation.} Dempster--Shafer fusion combines heterogeneous evidence with explicit ignorance and conflict, and exact Shapley values attribute aggregate task risk across the four attack categories (\S\ref{sec:quantrisk}).
\end{itemize}

\begin{table}[t]
\caption{The four perturbation types map to the four operative layers of an AgentWorld simulation. Each targets a distinct component of the agent's execution environment.}
\label{tab:perturbations}
\setlength{\tabcolsep}{2pt}
\fontsize{6.5}{8}\selectfont
\begin{tabular}{@{}llll@{}}
\toprule
\textbf{Type} & \textbf{Layer} & \textbf{Target / Effect} & \textbf{Example} \\
\midrule
State mutation & Data & Corrupts app state or & set account\_status \\
 & & persistent data fields & = LIMITED \\
Message injection & Comm. & Injects misleading or & inject ``cancel my \\
 & & conflicting user messages & request'' as the user \\
Action interception & Tool & Alters, blocks, or tampers & make refund return \\
 & & with tool call outputs & service error \\
System disruption & Infra. & Causes delays, timeouts, & timeout / partial-write \\
 & & or partial failures & on get\_cards \\
\bottomrule
\end{tabular}
\end{table}

\begin{figure*}[t]
    \centering
    \includegraphics[width=\textwidth]{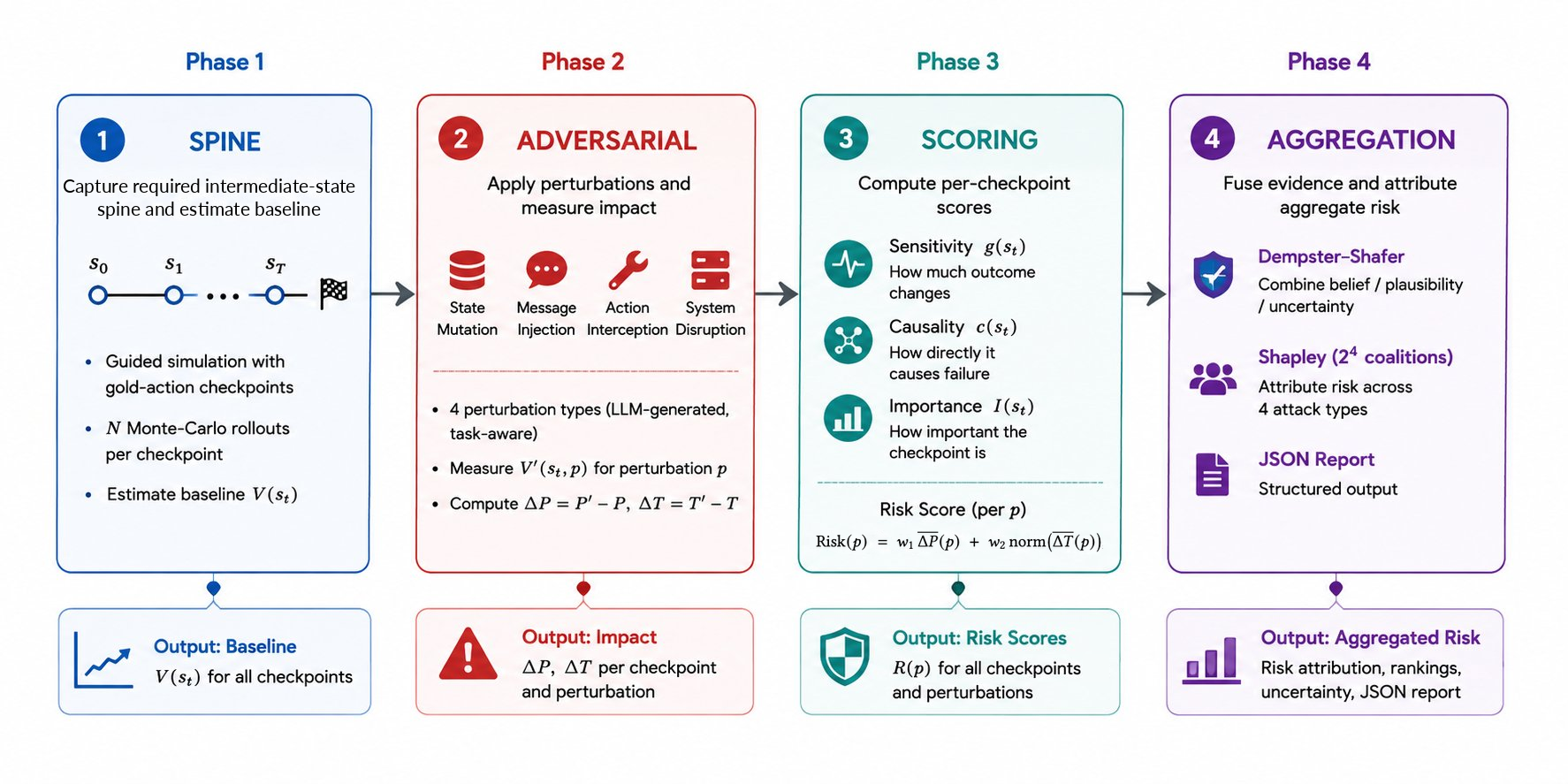}
    \caption{Risk Analyser: four-phase pipeline. Input: a completed simulation trajectory $\tau = (s_0, a_0, s_1, a_1, \ldots, s_T)$ and task goal $G$. Phase~1 (Spine): captures the required intermediate-state spine and estimates the baseline value function $V(s_t)$ from $N$ Monte-Carlo rollouts at each checkpoint. Phase~2 (Adversarial Perturbations): applies four types of task-aware perturbations and measures the perturbed value $V'(s_t, p)$ to compute $\Delta P$ and $\Delta T$. Phase~3 (Scoring): computes per-checkpoint sensitivity $g(s_t)$, causality $c(s_t)$, and importance $I(s_t)$, yielding a per-perturbation Risk Score. Phase~4 (Aggregation): fuses evidence via Dempster--Shafer theory and attributes aggregate risk across the four attack categories via exact Shapley values.}
    \label{fig:riskpipeline}
\end{figure*}

\begin{sloppypar}
\textbf{Spine and checkpoint/branch.} A single guided \emph{spine} simulation runs the task to completion, capturing the full simulation state (agents, apps, messages, action log) at every required intermediate state using AgentWorld's checkpoint primitives.
From each captured state~$s_t$ we branch~$N$ independent Monte-Carlo rollouts (default $N\!=\!8$) that diverge from the spine and execute autonomously to task termination. The spine is retried up to three times if intermediate-state coverage is incomplete.
Unperturbed rollouts give the baseline value~$V(s_t)$; perturbed rollouts give $V'(s_t, p)$ for each perturbation~$p$.
The architecture (Figure~\ref{fig:riskpipeline}) is a four-phase pipeline---spine~$\to$~adversarial rollouts~$\to$~scoring~$\to$~aggregation---that stacks above AgentWorld's evaluation phase.
\end{sloppypar}

\section{Evaluation Framework}
\label{sec:eval}

After a simulation completes, AgentWorld evaluates it through two independent scoring layers: a \textit{task-level} layer that checks whether the agent accomplished its goal, and a \textit{behavioural} layer that scores the quality of each individual message.

\subsection{Measuring Consistency: The pass$^k$ Metric}

A common mistake in agent testing is running a task once, seeing it succeed, and declaring it ready. But what if it fails three out of five times? Following $\tau$-bench~\cite{yao2024tau}, AgentWorld uses the pass$^k$ metric, which asks a stricter question: ``If we run this task $k$ times, what is the probability that the agent succeeds \textit{every single time}?''
\begin{equation}
\text{pass}^k = \frac{\binom{c}{k}}{\binom{n}{k}}
\end{equation}
\begin{sloppypar}
where $n$ is the total number of trials and $c$ is the number of successes. An agent with an 80\% success rate (8/10 trials) sounds production-ready, but $\text{pass}^8\!=\!2.2\%$---meaning there is only a 2.2\% chance that 8~users in a row all have a good experience. By contrast, the optimistic pass@$k$ would report 100\% at $k\!=\!8$. This is the gap between ``works in a demo'' and ``works in production.''
\end{sloppypar}

\subsection{Diagnostic Layers Beyond Pass/Fail}
\label{sec:faults}

A binary pass/fail label is rarely actionable. AgentWorld layers four diagnostic mechanisms on top of pass$^k$:

\textit{(1) Structured fault classification.} Every failure is tagged along two axes---\textit{who} was responsible (Agent, Environment, Task definition) and \textit{what} went wrong (wrong action, wrong parameters, missing action, policy violation, missing confirmation, reasoning error). This turns ``the agent failed'' into ``62\% of failures are Agent / Wrong-parameters,'' pointing teams at parameter extraction rather than the base model.

\textit{(2) Partial-credit scoring.} Instead of pass-or-nothing, AgentWorld awards partial credit $\text{credit} = 0.5 \cdot s_{\text{completed}}/s_{\text{total}} + 0.5 \cdot f_{\text{correct}}/f_{\text{total}}$ where $s$ is steps completed and $f$ is correct state fields. A task failing at step 9/10 with 4/5 fields correct scores 0.85, distinguishing ``almost worked'' from ``failed immediately.''

\textit{(3) Policy compliance.} Teams define safety rules declaratively (e.g., \textit{``confirm before executing transfers $>$ \$500''}) with conditions, required behaviours, and severity levels. AgentWorld ships with pre-built policy sets for payment (8 rules) and shopping (9 rules).

\textit{(4) Per-message behavioural scoring.} Every individual message is scored by LLM judges rating \textit{persona adherence}, \textit{coherence}, \textit{relevance}, \textit{consistency}, plus heuristic checks for length and keyword safety. These scores decide which conversations enter the training-data export and enable persona-differentiated analysis (Section~\ref{sec:experiments}).

\subsection{Quantifying Coordination Cost}

For dual-control tasks (Section~\ref{sec:apps}, Figure~\ref{fig:handoff}), a single number summarises how much performance degrades when the agent must coordinate via handoffs rather than acting alone. We adopt the dual-control architecture of $\tau^2$-bench~\cite{yao2025tau2}: the service agent controls backend tools while the user controls a separate device app. The \textit{coordination overhead} is the gap between solo mode (agent gets full tool access and acts alone) and dual mode (agent must instruct, the user must act):
\begin{equation}
\Delta_{\text{coord}} = \text{pass}^1_{\text{solo}} - \text{pass}^1_{\text{dual}}
\end{equation}
A large $\Delta_{\text{coord}}$ tells teams the failure is in conversational design (handoff phrasing, confirmation logic) rather than capability---a fix that requires prompt and policy changes, not a stronger base model.

\subsection{Quantifying Adversarial Risk}
\label{sec:quantrisk}

While pass$^k$ quantifies reliability under aleatoric noise, the Risk Analyser quantifies reliability under adversarial perturbation. This subsection formalises the scoring pipeline (Figure~\ref{fig:riskpipeline}, Phases~3--4).

\textbf{Value function and counterfactual signals.} Let $s_t$ denote the simulation state captured at the $t$-th required intermediate state. The success-probability value function and its perturbed counterpart are estimated by Monte-Carlo branching:
\begin{equation}
V(s_t) = P(\text{success} \mid s_t) \approx \frac{1}{N} \sum_{i=1}^{N} \mathbf{1}\{\rho_i\},
\label{eq:value}
\end{equation}
$V'(s_t, p)$ analogously under perturbation $p$, where $\rho_i \in \{0, 1\}$ is the success-indicator function of the $i$-th rollout (1 if the rollout reaches the task's success conditions, 0 otherwise). The two derived per-checkpoint quantities are the failure-probability gap and the extra-turn cost:
\begin{multline}
\Delta P(s_t, p) = V(s_t) - V'(s_t, p),\\
\Delta T(s_t, p) = \mathbb{E}[T'] - \mathbb{E}[T].
\label{eq:deltas}
\end{multline}

\textbf{Sensitivity and causal scores.} Every checkpoint is scored along two complementary axes:
\begin{multline}
g(s_t) = \frac{1}{D} \sum_{d=1}^{D} \frac{\bigl|V(s_t) - V(\text{perturb}_d(s_t))\bigr|}{|\Delta_d|},
\label{eq:sensitivity}
\end{multline}
\begin{multline}
c(s_t) = \log P(y^\star \mid \tau_{1:T}) \\
- \log P(y^\star \mid \tau^{[s_t \leftarrow \text{corrupt}]}_{1:T}),
\label{eq:causal}
\end{multline}
where $g(s_t)$ is a finite-difference discrete sensitivity averaged over~$D$ state dimensions perturbed at~$s_t$---capturing the local response of~$V$ to small state changes---while $c(s_t)$ is a counterfactual log-likelihood drop induced by corrupting~$s_t$ within the trajectory~$\tau$ relative to the eventual outcome~$y^\star$---capturing path-level causal necessity of~$s_t$.
The two are blended into a single per-checkpoint importance:
\begin{equation}
I(s_t) = \alpha\, \tilde{g}(s_t) + (1 - \alpha)\, \tilde{c}(s_t), \quad \alpha \in [0, 1],
\label{eq:importance}
\end{equation}
\begin{sloppypar}
with $\tilde{g}, \tilde{c}$ min-max normalised to $[0, 1]$ and $\alpha\!=\!0.5$ as the default.
The form is inspired by counterfactual step-importance methods from chain-of-thought reasoning~\cite{madaan2022}: $g$~captures local geometry, $c$~captures global causal structure, and the convex combination spans the spectrum.
In the experiments reported here we use the default single-dimension perturbation ($D\!=\!1$), under which $c(s_t)$ reduces to~$g(s_t)$ and $I(s_t)$ collapses to the sensitivity term; the $\alpha$-blend above is the general form, exercised only when multi-dimensional perturbations make $g$ and~$c$ diverge.
\end{sloppypar}

\textbf{Risk Score.} Per-perturbation risk averages over checkpoints and combines failure-probability and delay terms:
\begin{equation}
\text{Risk}(p) = w_1 \overline{\Delta P(p)} + w_2\, \text{norm}(\overline{\Delta T(p)}),
\label{eq:riskscore}
\end{equation}
where $\overline{\cdot}$ is the mean over checkpoints, $\mathrm{norm}(\cdot)$ is min-max normalisation across the perturbation set, and $w_1, w_2$ are profile-configurable (default $w_1\!=\!0.7$, $w_2\!=\!0.3$).
Three weighting schemes for converting~$I(s_t)$ to per-checkpoint weights---baseline-plus-boost ($w_t\!=\!1\!+\!\beta I_t$), min-max bounded, and softmax-with-temperature---are supported, corresponding to conservative, balanced, and aggressive risk attitudes.

\textbf{Dempster--Shafer evidence fusion.}
A single checkpoint accumulates risk evidence from heterogeneous sources of varying confidence; with small~$N$ this evidence carries genuine epistemic uncertainty rather than just noise.
Define the frame of discernment $\Theta = \{\textrm{risk},\allowbreak \neg\textrm{risk}\}$.
Each source supplies a mass function~$m(\cdot)$ over $\bigl\{\{\textrm{risk}\},\allowbreak \{\neg\textrm{risk}\},\allowbreak \Theta\bigr\}$, with mass on~$\Theta$ representing ignorance as first-class~\cite{shafer1976}.
Table~\ref{tab:massfunctions} lists the four sources.
Dempster's normalised rule of combination produces a fused mass:
\begin{multline}
m_{12}(A) = \frac{1}{1 - k}\!\!\sum_{B \cap C = A}\!\! m_1(B)\,m_2(C),\\
k = \!\!\sum_{B \cap C = \varnothing}\!\! m_1(B)\,m_2(C),
\label{eq:dempster}
\end{multline}
where the conflict coefficient~$k$ is reported alongside the result and flags high-disagreement cases ($k > 0.7$).
Per-checkpoint Belief $\mathrm{Bel}(\textrm{risk})\allowbreak = m(\{\textrm{risk}\})$, Plausibility $\mathrm{Pl}(\textrm{risk})\allowbreak = 1 - m(\{\neg\textrm{risk}\})$, and Uncertainty $\mathrm{Pl} - \mathrm{Bel}$ are aggregated to task level via importance-weighted average:
\begin{equation}
\mathrm{Bel}_{\mathrm{task}} = \textstyle\sum_t \omega_t\,\mathrm{Bel}(s_t),\quad \omega_t = I(s_t)\big/\textstyle\sum_s I(s).
\end{equation}
Three choices distinguish this from textbook D--S\@: $\sqrt{N}$-scaled MC mass with one-sided informational sources (Table~\ref{tab:massfunctions}), importance-weighted task aggregation, and explicit conflict reporting---together preventing the dilution of strong evidence and the silent masking of source disagreement that flat-averaging baselines produce.

\begin{table}[t]
\caption{Mass functions per evidence source in Dempster--Shafer fusion. Each source assigns mass to three hypotheses: $m(\text{risk})$ = belief that the checkpoint is risky, $m(\neg\text{risk})$ = belief that it is safe, and $m(\Theta)$ = residual ignorance. One-sided sources contribute evidence for risk only, encoding that evidence of vulnerability is not evidence of safety. Notation: $V$ = baseline success probability; $\tilde{g}(s_t)$ = normalised sensitivity; $\tilde{c}(s_t)$ = normalised causal score; $\Delta P$ = failure-probability gap.}
\label{tab:massfunctions}
\setlength{\tabcolsep}{3pt}
\fontsize{6.5}{8}\selectfont
\begin{tabular}{@{}lccc@{}}
\toprule
\textbf{Source} & $m(\text{risk})$ & $m(\neg\text{risk})$ & $m(\Theta)$ \\
\midrule
MC rollouts ($\text{conf} = 1 - 1/\!\sqrt{N}$) & $(1\!-\!V)\,\text{conf}$ & $V\,\text{conf}$ & $1\!-\!\text{conf}$ \\
Sensitivity (one-sided) & $\tilde{g}(s_t)$ & 0 & $1 - \tilde{g}(s_t)$ \\
Causal (one-sided) & $\tilde{c}(s_t)$ & 0 & $1 - \tilde{c}(s_t)$ \\
Max adversarial $\Delta P$ (one-sided) & $\max_p \Delta P$ & 0 & $1 - \max_p \Delta P$ \\
\bottomrule
\end{tabular}
\end{table}

\textbf{Shapley attribution across attack categories.}
To attribute aggregate task risk across the four attack types $\mathcal{N} = \{\textrm{state},\allowbreak \textrm{message},\allowbreak \textrm{action},\allowbreak \textrm{system}\}$, we compute exact Shapley values~\cite{shapley1953}:
\begin{multline}
\phi_i = \!\!\sum_{S \subseteq \mathcal{N} \setminus \{i\}}\!\! \frac{|S|!\,(|\mathcal{N}|\!-\!|S|\!-\!1)!}{|\mathcal{N}|!}\\
\times\bigl[v(S \cup \{i\}) - v(S)\bigr],
\label{eq:shapley}
\end{multline}
with coalition value $v(S) = \min\!\bigl(1,\allowbreak \sum_{i \in S} \overline{\Delta P_i}\bigr)$.
With $|\mathcal{N}| = 4$, exact computation over all~$2^4 = 16$ coalitions is tractable; the four Shapley axioms (efficiency, symmetry, null player, additivity) guarantee a fair, principled attribution capturing interaction effects between categories.
Attack-graph kill-chain analysis~\cite{sheyner2002} is left to future work, as the current single-attack-dominance regime makes multi-step chains uninformative.

\subsection{Evaluate-to-Train Closed Loop}
\label{sec:export}

The export pipeline ties directly to the evaluation layer: a minimum behavioural-score threshold gates which conversations enter the training set, DPO pairs are auto-constructed from high- and low-scoring responses to the same stimulus, and every exported example carries full provenance (persona config, per-evaluator scores, evaluator model, timestamp). Three redaction profiles---None, Basic, Strict---govern information sharing. Because each conversation is grounded in a specific OCEAN trait vector, the exported dataset has structural diversity in directness, emotional valence, and engagement style that naive prompt perturbation cannot achieve.

\subsection{Personality-Stratified Model Comparison}
\label{sec:modelcomp}

A critical use case is evaluating whether fine-tuning improves performance \textit{across the full personality spectrum}.
The \texttt{Model\-Comparison\-Runner} runs a baseline and one or more fine-tuned checkpoints through identical simulation packs, producing comparison reports stratified by OCEAN quadrant.
This enables \textit{personality-aware early stopping} (best pass$^k$ on the hardest quadrant rather than lowest aggregate loss), \textit{regression detection} (does epoch~$N$ degrade cooperative-user performance while improving adversarial?), and \textit{fault-category tracking} (e.g., \texttt{WRONG\_PARAMETERS}: $18\%\!\rightarrow\!12\%\!\rightarrow\!7\%$ across epochs).
The \texttt{Experiment\-Runner} adds A/B testing with paired significance tests and Cohen's~$d$ effect sizes.

\section{Experiments}
\label{sec:experiments}

\subsection{Experiment 1: Persona-Panel Evaluation of an Analytics Agent}
\label{sec:exp1}

To demonstrate AgentWorld's diagnostic capabilities on a real agent, we evaluated a production \textit{conversational analytics agent}---an LLM-based system that provides natural-language access to business KPIs across geographies, segments, and products. The agent was integrated into AgentWorld with zero code changes.

\textbf{Setup.} Ten analyst personas were configured, each combining an OCEAN trait vector with a distinct role and query domain (e.g., campaign analysis, regional market deep-dives, executive synthesis, data quality auditing, competitive intelligence). Each persona ran 3 multi-turn conversation exchanges against the analytics agent, producing 60 messages (30 from the agent, 30 from personas) and 240 evaluator judgments (4 evaluators $\times$ 60 messages). An LLM judge scored coherence and relevance; heuristic evaluators scored length and keyword safety. All \textbf{30/30 agent calls succeeded} with no fallbacks. Average response length showed a \textbf{7.2$\times$ persona-driven range}---from 244 chars for the most terse persona to 1762 chars for the most verbose---direct evidence that persona configuration shapes conversation \textit{shape}, not just content.

\textit{Scope note:} because each persona here varies role and query domain jointly with its trait vector, this experiment is a persona-panel/domain-coverage demonstration; it is not designed to isolate the causal effect of personality alone. The controlled personality-vs-task analysis---identical task, OCEAN variant as the only manipulated factor---is provided by Experiment~2 (\S\ref{sec:exp2}).

\textbf{Aggregate results.} At the aggregate level, the agent appeared to perform well (Table~\ref{tab:exp1_agg}). A single-number evaluation would report coherence of 0.79 and relevance of 0.75 as an acceptable agent. The per-persona breakdown tells a different story.

\begin{table}[h!]
\caption{Experiment~1: Evaluation of an analytics agent across 10 OCEAN-parameterised personas (60 messages, 240 evaluator judgments). (a)~Aggregate scores across all messages. (b)~Relevance by persona panel (agent responses only), ranked best to worst---aggregate relevance of 0.75 conceals a range from 0.88 to 0.61.}
\label{tab:exp1_agg}
\small
\begin{subtable}[t]{\columnwidth}
\caption{Aggregate scores}
\begin{tabular}{@{}lcccc@{}}
\toprule
\textbf{Evaluator} & \textbf{Mean} & \textbf{Min} & \textbf{Max} & \textbf{$n$} \\
\midrule
Coherence & 0.793 & 0.45 & 0.95 & 60 \\
Relevance & 0.749 & 0.30 & 0.95 & 60 \\
Length check & 0.990 & 0.70 & 1.00 & 60 \\
Keyword filter & 1.000 & 1.00 & 1.00 & 60 \\
\bottomrule
\end{tabular}
\end{subtable}

\vspace{0.3cm}

\begin{subtable}[t]{\columnwidth}
\caption{Relevance by persona panel}
\begin{tabular}{@{}clcc@{}}
\toprule
\textbf{Rank} & \textbf{Persona Role} & \textbf{Rel.} & \textbf{Coh.} \\
\midrule
1 & Campaign Analyst & 0.88 & 0.78 \\
2 & Competitive Intelligence & 0.83 & 0.56 \\
3 & Regional Market Specialist & 0.79 & 0.81 \\
4 & Data Quality Auditor & 0.78 & 0.85 \\
5 & Performance Analyst & 0.77 & 0.78 \\
6 & Forecasting Analyst & 0.76 & 0.70 \\
7 & Engagement Analyst & 0.73 & 0.89 \\
8 & Product Manager & 0.68 & 0.86 \\
9 & Growth Analyst & 0.67 & 0.84 \\
10 & Executive Synthesis & 0.61 & 0.88 \\
\bottomrule
\end{tabular}
\end{subtable}
\end{table}

\textbf{Per-persona breakdown.} Breaking relevance out by persona panel reveals a clear domain-gap pattern (Table~\ref{tab:exp1_agg}b)---aggregate relevance of 0.75 spans 0.88 (Campaign Analyst) to 0.61 (Executive Synthesis). Because role and domain co-vary with personality here, this spread reflects combined role/domain/personality difficulty rather than personality in isolation; Experiment~2 isolates the personality factor on a fixed task.

\textbf{Failure patterns discovered.} Structured evaluation revealed three systematic failures invisible to aggregate scoring. (1)~\textit{Response truncation} (5/30 responses, coherence 0.55--0.62): the agent's output was cut off mid-sentence due to a silent token-limit on long analytical responses. (2)~\textit{Contextual drift in multi-turn} (3 instances, worst case 0.30 relevance): the agent answered the literal question but failed to connect its response to the ongoing thread---the judge noted it ``introduces a new subtopic without connecting it to the previous discussion points.'' (3)~\textit{Cross-domain data leakage} (1 instance, 0.60 relevance): when a persona focused on one geographic market asked about demographics, the agent cited data from a different region.

\subsection{Experiment 2: Personality-Stratified Evaluation of a Customer-Support Agent}
\label{sec:exp2}

While Experiment~1 demonstrated persona-panel evaluation on an analytics agent, Experiment~2 scales the evaluation to a full personality $\times$ task matrix---19 simulations across 5 customer-support tasks and 4 OCEAN persona variants---and then stress-tests the same tasks under adversarial perturbation via the Risk Analyser.

\begin{sloppypar}
\textbf{Setup.} We evaluated a production customer-support agent (\texttt{gpt-5.2}) with access to payment, account, and transaction tools against four OCEAN persona variants---\textit{Impatient} ($t_E\!=\!0.9,\allowbreak t_N\!=\!0.8$), \textit{Analytical} ($t_C\!=\!0.9$), \textit{Aggressive} ($t_A\!=\!0.1,\allowbreak t_N\!=\!0.8$), and \textit{Anxious} ($t_N\!=\!0.9,\allowbreak t_E\!=\!0.2$)---simulated by \texttt{gpt-5-mini}.
Five tasks of increasing complexity were selected: fetch latest transactions (read-only), check refund eligibility (status-check), cancel pending payments (state-mutating), complete payment for pending bills (state-mutating), and withdraw funds to a bank account (state-mutating with verification).
LLM judges scored every message on five criteria: persona adherence, relevance, coherence, consistency, and task completeness. Table~\ref{tab:exp2} reports all 19~results. Judge scores come from one simulation per cell (Paper Sim Pack, May 2025); pass$^k$ columns come from a separate reliability study ($n\!=\!16$ reruns per cell) recording goal-state completion only.
\end{sloppypar}

\begin{table*}[!t]
\caption{Experiment~2: 19 (task$\times$persona) cells on a customer-support agent (mesh topology). Left block: one simulation per cell (agent \texttt{gpt-5.2}, customer \texttt{gpt-5-mini}); LLM judges score five criteria per message; Judge = PASS if aggregated overall score $\geq$ 0.70. Right block: $n\!=\!16$ independent reruns per cell (both roles \texttt{gpt-5.2}); pass$^k = \binom{c}{k}/\binom{n}{k}$ on goal-state completion (judge scores not re-aggregated over reruns). Steps = turns/budget; cs/ms = customer\_signal / max\_steps. $\dagger$Judge and pass$^k$ diverge.}
\label{tab:exp2}
\setlength{\tabcolsep}{3pt}
\fontsize{6}{7.5}\selectfont
\begin{tabular}{@{}ll c ccccc cc cccc@{}}
\toprule
 & & & \multicolumn{5}{c}{\textbf{Single run ($n\!=\!1$; judge)}} & & & \multicolumn{4}{c}{\textbf{Reliability ($n\!=\!16$; goal state)}} \\
\cmidrule(lr){4-8} \cmidrule(lr){11-14}
\textbf{Task} & \textbf{Persona} & \textbf{Steps} & \textbf{Overall} & \textbf{P.Adh} & \textbf{Rel} & \textbf{Coh} & \textbf{Con} & \textbf{Compl} & \textbf{Judge} & \textbf{pass$^1$} & \textbf{pass$^2$} & \textbf{pass$^4$} & \textbf{pass$^8$} \\
\midrule
Complete Payment  & Impatient  & 3/15\,cs  & 0.906 & 0.742 & 0.967 & 0.938 & 0.933 & 0.950 & PASS & 1.00 & 1.00 & 1.00 & 1.00 \\
for Pending Bills & Analytical & 4/15\,cs  & 0.875 & 0.857 & 0.723 & 0.909 & 0.939 & 0.950 & PASS & 1.00 & 1.00 & 1.00 & 1.00 \\
                  & Aggressive & 3/15\,cs  & 0.846 & 0.680 & 0.697 & 0.933 & 0.970 & 0.950 & PASS & 1.00 & 1.00 & 1.00 & 1.00 \\
                  & Anxious    & 15/15\,ms & 0.761 & 0.842 & 0.780 & 0.868 & 0.892 & 0.420 & PASS & 1.00 & 1.00 & 1.00 & 1.00 \\
\midrule
Withdraw          & Aggressive & 2/15\,cs  & 0.793 & 0.698 & 0.743 & 0.955 & 0.950 & 0.620 & PASS & 1.00 & 1.00 & 1.00 & 1.00 \\
to Bank           & Impatient  & 15/15\,ms & 0.858 & 0.830 & 0.876 & 0.936 & 0.928 & 0.720 & PASS & 1.00 & 1.00 & 1.00 & 1.00 \\
                  & Analytical & 15/15\,ms & 0.824 & 0.869 & 0.793 & 0.863 & 0.875 & 0.720 & PASS & 1.00 & 1.00 & 1.00 & 1.00 \\
                  & Anxious    & 15/15\,ms & 0.777 & 0.834 & 0.855 & 0.937 & 0.840 & 0.420 & PASS & 1.00 & 1.00 & 1.00 & 1.00 \\
\midrule
Cancel Pending    & Impatient  & 10/10\,ms & 0.836 & 0.824 & 0.884 & 0.912 & 0.940 & 0.620 & PASS & 1.00 & 1.00 & 1.00 & 1.00 \\
Payments          & Anxious    & 10/10\,ms & 0.749 & 0.866 & 0.813 & 0.929 & 0.936 & 0.200 & PASS & 1.00 & 1.00 & 1.00 & 1.00 \\
                  & Analytical & 10/10\,ms & 0.742 & 0.894 & 0.867 & 0.923 & 0.827 & 0.200 & PASS & 0.94 & 0.88 & 0.75 & 0.50 \\
                  & Aggressive & 10/10\,ms & 0.714 & 0.814 & 0.738 & 0.927 & 0.813 & 0.280 & PASS & 0.94 & 0.88 & 0.75 & 0.50 \\
\midrule
Check Refund      & Aggressive & 10/10\,ms & 0.776 & 0.768 & 0.861 & 0.887 & 0.742 & 0.620 & PASS$^\dagger$ & 0.88 & 0.76 & 0.55 & 0.23 \\
Eligibility       & Anxious    & 10/10\,ms & 0.769 & 0.845 & 0.856 & 0.908 & 0.816 & 0.420 & PASS$^\dagger$ & 0.88 & 0.76 & 0.55 & 0.23 \\
                  & Analytical & 10/10\,ms & 0.758 & 0.904 & 0.804 & 0.851 & 0.792 & 0.440 & PASS$^\dagger$ & 0.94 & 0.88 & 0.75 & 0.50 \\
\midrule
Fetch Latest      & Analytical & 10/10\,ms & 0.755 & 0.861 & 0.782 & 0.789 & 0.822 & 0.520 & PASS & 0.94 & 0.88 & 0.75 & 0.50 \\
Transactions      & Anxious    & 10/10\,ms & 0.724 & 0.876 & 0.829 & 0.839 & 0.875 & 0.200 & PASS & 1.00 & 1.00 & 1.00 & 1.00 \\
                  & Impatient  & 10/10\,ms & 0.685 & 0.767 & 0.764 & 0.816 & 0.879 & 0.200 & FAIL$^\dagger$ & 1.00 & 1.00 & 1.00 & 1.00 \\
                  & Aggressive & 10/10\,cs & 0.678 & 0.653 & 0.861 & 0.805 & 0.872 & 0.200 & FAIL$^\dagger$ & 0.88 & 0.76 & 0.55 & 0.23 \\
\midrule
\multicolumn{2}{@{}l}{\textbf{Aggregate (19 sims)}} & & \textbf{0.779} & \textbf{0.812} & \textbf{0.815} & \textbf{0.891} & \textbf{0.876} & \textbf{0.508} & \multicolumn{5}{l}{17/19 judge pass per-cell; not pooled} \\
\bottomrule
\end{tabular}

\vspace{2pt}
{\fontsize{5.5}{7}\selectfont \textit{Notes:} Judge scores from Paper Sim Pack (May 2025; one run per cell). pass$^k$ from a separate reliability rerun; Check Refund / Fetch Latest used semantically equivalent evaluation DB task configs.}
\end{table*}

\textbf{Results.} Under LLM-judge criteria, 17 of 19 single-run simulations passed and 2 failed (89.5\%). The aggregate scores---coherence 0.891, consistency 0.876, relevance 0.815, persona adherence 0.812, completeness 0.508---reveal that the agent communicates well but often fails to complete tasks fully in the judge's assessment.

\textbf{Finding 1: Personality predicts judge-assessed failure.} Both judge failures occurred on Fetch Latest Transactions with the Impatient (Overall 0.685) and Aggressive (0.678) personas---driven by low completeness (0.200) despite coherent dialogue. The Analytical persona passed at 0.755 on the same task. A single-persona evaluation using only the Analytical profile would have reported 100\% judge pass. Judge failure here reflects conversation quality on one run, not necessarily goal-state unreliability (e.g., Fetch/Impatient: judge FAIL but pass$^1$=1.00 over 16 goal-state reruns).

\textbf{Finding 2: Large persona-driven quality gaps.} Within-task LLM-judge score ranges reached 0.145 points (Complete Payment: Impatient 0.906 vs.\ Anxious 0.761) and 0.122 points (Cancel Payments: Impatient 0.836 vs.\ Aggressive 0.714). These gaps are masked by the 0.779 aggregate and identify which persona--task pairs need targeted improvement.

\textbf{Finding 3: Completeness is the bottleneck, driven by step-budget exhaustion.} Completeness averaged only 0.508 while all other criteria exceeded 0.81. The Steps column in Table~\ref{tab:exp2} reveals the mechanism: 14 of 19 simulations hit their step budget (ms = max\_steps), while only 5 terminated via natural conversation end (cs = customer\_signal). The contrast is sharpest within a single task---Complete Payment finished in just 3 steps for the Impatient persona (0.950 completeness) but consumed all 15 steps for the Anxious persona (0.420 completeness), whose repeated clarification requests exhausted the budget before all goal conditions were met. Token usage mirrored this pattern, ranging from $\sim$23K (Withdraw, Aggressive, 2 steps) to $\sim$187K (Withdraw, Anxious, 15 steps). Low completeness on a single judge-scored run does not always imply poor goal-state reliability on repeat (see pass$^k$ columns).

\textbf{Finding 4: pass$^k$ exposes reliability gaps that single judge runs hide.} Rerunning every cell $n\!=\!16$ times on goal-state success (last four columns of Table~\ref{tab:exp2}), Complete Payment and Withdraw to Bank stay perfectly reliable (pass$^k$=1.00), but other tasks erode as $k$ grows---Check Refund Eligibility drops from pass$^1$=0.88--0.94 to pass$^8$=0.23--0.50 despite judge PASS on the original run. High judge scores on a single run thus mask cells where the agent cannot sustain repeated task completion---a demo-vs-production gap the Judge column alone cannot reveal.

\textbf{Forward pointer to Experiment~3.} Experiment~3 (\S\ref{sec:exp3}) stress-tests the same task family under adversarial perturbation to reveal trajectory-level brittleness that behavioural scoring cannot see.

\subsection{Experiment 3: Adversarial Stress-Test via the Risk Analyser}
\label{sec:exp3}

\textbf{Setup.} We apply the Risk Analyser (\S\ref{sec:riskanalyser}, \S\ref{sec:quantrisk}) to five customer-support tasks---three drawn from Experiment~2 (Check Refund Eligibility, Cancel Pending Payments, Fetch Latest Transactions) plus two additional probes (File an INR Dispute, Add a New Card) extending coverage beyond Experiment~2's set. The agent and topology match Experiment~2 (\texttt{gpt-5.2}, mesh). Per task we run $N\!=\!8$ baseline and $N\!=\!8$ adversarial Monte-Carlo rollouts at every required intermediate state, with $|P|\!=\!8$ task-aware perturbations (2 per type). This yields several hundred Monte-Carlo rollouts per task ($N\!=\!8$ at each required intermediate state, plus $N\!=\!8$ under each of the $|P|\!=\!8$ perturbations) and on the order of a thousand across the five tasks; the per-checkpoint sample ($N\!=\!8$) is deliberately small, which we revisit as an uncertainty caveat below. Scoring uses defaults: $\alpha\!=\!0.5$, $w_1\!=\!0.7$, $w_2\!=\!0.3$.

\begin{table}[t]
\caption{Risk Analyser reveals trajectory-level brittleness invisible to pass-rate testing. $V_{\min}$ = lowest baseline success probability across checkpoints ($<$1.0 = pre-existing brittleness); top $\Delta P$ = largest failure-probability gap (attack type in parentheses); task Bel = D--S fused belief (lower bound on risk); Shapley \% = risk attribution to State / Message / Action / Infra attacks. Config: $N$=8/8, $|P|$=8.}
\label{tab:risk}
\setlength{\tabcolsep}{2.5pt}
\fontsize{6.5}{8}\selectfont
\begin{tabular}{@{}l c c c c c@{}}
\toprule
 & pass & & top $\Delta P$ & task & Shapley \% \\
\textbf{Task} & \% & $V_{\min}$ & (type) & Bel & (S/M/A/I) \\
\midrule
Fetch Latest Transactions & 50 & 1.00 & .333 (sys.) & .900 & 0/0/53/47 \\
Check Refund Eligibility & 100 & .375 & .458 (state) & .925 & 28/12/22/38 \\
Cancel Pending Payments & 100 & 1.00 & .500 (act.) & 1.00 & 0/0/50/50 \\
File an INR Dispute & 100 & 1.00 & .800 (act.) & 1.00 & 3/5/31/61 \\
Add a New Card & 100 & 1.00 & .333 (sys.) & 1.00 & 32/4/32/32 \\
\midrule
\multicolumn{5}{@{}l}{Mean Shapley:} & 12/4/38/46 \\
\bottomrule
\end{tabular}
\end{table}

\begin{table}[t]
\caption{D--S worked example at checkpoint $t\!=\!1$ (the first required intermediate state) of Check Refund Eligibility. Baseline $V(s_t)\!=\!0.375$ ($N\!=\!8$, 3/8 succeed)---pre-existing brittleness invisible to behavioural scoring. Dempster's rule fuses three active sources to Bel=0.727, Pl=0.889, Uncertainty=0.162, conflict $k$=0.062.}
\label{tab:ds_example}
\setlength{\tabcolsep}{3pt}
\fontsize{6.5}{8}\selectfont
\begin{tabular}{@{}lccc@{}}
\toprule
\textbf{Evidence Source} & $m(\text{risk})$ & $m(\neg\text{risk})$ & $m(\Theta)$ \\
 & (risky) & (safe) & (uncertain) \\
\midrule
MC rollouts ($N\!=\!8$, conf=0.646) & 0.404 & 0.242 & 0.354 \\
Sensitivity ($\tilde{g}\!=\!0.375$, one-sided) & 0.375 & 0 & 0.625 \\
Causal & \multicolumn{3}{c}{collapsed into sensitivity ($\tilde{c}\!=\!\tilde{g}$)} \\
Max $\Delta P$ ($=\!0.375$, one-sided) & 0.375 & 0 & 0.625 \\
\midrule
\textbf{Fused result} & \textbf{0.727} & \textbf{0.111} & \textbf{0.162} \\
\bottomrule
\multicolumn{4}{@{}l}{\fontsize{5.5}{7}\selectfont Bel=0.727, Pl=0.889, Uncertainty=0.162, conflict $k$=0.062} \\
\end{tabular}
\end{table}

\begin{sloppypar}
\textbf{Aggregate results.} Table~\ref{tab:risk} contrasts each task's behavioural pass-rate against four Risk Analyser metrics: $V_{\min}$, top~$\Delta P$ and its attack type, task-level Belief, and per-attack-category Shapley share.
Table~\ref{tab:ds_example} walks through a single D--S fusion at the first checkpoint of Check Refund Eligibility, where three active evidence sources (the causal source collapses into sensitivity under the default single-dim perturbation) fuse to $\mathrm{Bel}\!=\!0.727$, $\mathrm{Pl}\!=\!0.889$, conflict $k\!=\!0.062$.
All Risk Analyser estimates derive from $N\!=\!8$ rollouts and carry correspondingly wide uncertainty: the headline $V_{\min}\!=\!0.375$ has a 95\% Wilson interval of $[0.14, 0.69]$, so reported values should be read as point estimates pending higher-$N$ confirmation. Four findings:
\end{sloppypar}

\textit{(1) Pass-rates mask trajectory brittleness.} Check Refund Eligibility achieved 100\% behavioural pass yet $V_{\min}\!=\!0.375$ at its first required intermediate state (only 3/8 unperturbed rollouts succeed). State-mutation drives the $\Delta P\!=\!0.458$ peak; D--S fuses to Bel=0.727, confirming genuine risk (Table~\ref{tab:ds_example}).

\begin{sloppypar}
\textit{(2) Tool/infra layers dominate risk.} Shapley attributes 46\% to system-disruption and 38\% to action-interception, versus only 12\% state and 4\% message.
The most damaging perturbation---action-interception on File an INR Dispute ($\Delta P\!=\!0.800$)---targets the layer that behavioural scoring cannot observe.
\end{sloppypar}

\textit{(3) Two complementary views.} Behavioural evaluation (Experiment~2) identifies which personas and tasks fail; the Risk Analyser (Experiment~3) identifies which trajectory steps and system layers are brittle. Together they provide a two-dimensional reliability map.

\textit{(4) Fusion is not averaging.} On the Check Refund checkpoint, flat-averaging the same evidence sources yields a risk mass of 0.385, whereas Dempster--Shafer fusion concentrates the concordant evidence to Bel=0.727 ($k$=0.062)---nearly double---isolating the marginal contribution of the D--S layer over naive aggregation.

\section{Discussion and Limitations}

\textbf{Limitations.} The persona-to-behaviour mapping relies on prompt engineering without validated calibration against human studies: we do not yet verify that a simulated persona with a given OCEAN vector elicits the same agent behaviour a real user of that profile would, so the personas should be read as controlled, reproducible stress profiles rather than validated human proxies. Establishing that link requires a human-subjects study (matching simulated personas to real users on Big Five instruments and comparing elicited agent behaviour), which we leave to future work. Relatedly, our conclusions depend on LLM judges, which introduce their own biases; a cross-judge agreement analysis is needed to bound this dependence, and DPO auto-construction uses score differentials as a proxy for human preference. On the adversarial side, sensitivity $g(s_t)$ and causal $c(s_t)$ are numerically equivalent under the default single-dimension perturbation (multi-dimensional perturbations separate them); Dempster's rule degrades under high conflict ($k > 0.7$), which we surface but do not yet resolve; and attack-graph kill-chain analysis is deferred because current single-attack dominance makes multi-step chains uninformative.

\textbf{Closing.} AgentWorld unifies personality-driven simulation, reliability measurement, training-data generation, and adversarial risk analysis in one closed-loop pipeline---exposing IR failure modes that uniform testing misses and making personality-aware reliability evaluation practical for agentic IR.

\bibliographystyle{ACM-Reference-Format}

{\fontsize{7.5}{9}\selectfont

}

\end{document}